\documentclass[runningheads]{llncs}
\usepackage{graphicx}
\usepackage{amsmath,amssymb} 
\usepackage{color}

\begin{document}
\title{Hierarchical Video Frame Sequence Representation with Deep Convolutional Graph Network} 

\titlerunning{Hierarchical Sequence Representation with Graph Network}
%

\author{Feng Mao \orcidID{0000-0001-6171-3168} \and
Xiang Wu  \orcidID{0000-0003-2698-2156} \and
Hui Xue \and
Rong Zhang}
%
\authorrunning{Feng Mao and Xiang Wu}
%

\institute{Alibaba Group, Hangzhou, China \\
\email{\{maofeng.mf,weiyi.wx\}@alibaba-inc.com}}
\maketitle              
\begin{abstract}
High accuracy video label prediction (classification) models are attributed to large scale data. These data could be frame feature sequences extracted by a pre-trained convolutional-neural-network, which promote the efficiency for creating models. Unsupervised solutions such as feature average pooling, as a simple label-independent parameter-free based method, has limited ability to represent the video. While the supervised methods, like RNN, can greatly improve the recognition accuracy. However, the video length is usually long, and there are hierarchical relationships between frames across events in the video, the performance of RNN based models are decreased. In this paper, we proposes a novel video classification method based on a deep convolutional graph neural network(DCGN). The proposed method utilize the characteristics of the hierarchical structure of the video, and performed multi-level feature extraction on the video frame sequence through the graph network, obtained a video representation reflecting the event semantics hierarchically. We test our model on YouTube-8M Large-Scale Video Understanding dataset, and the result outperforms RNN based benchmarks.

\keywords{Video Classification \and Sequence Representation \and Graph Neural Network \and Deep Convolutional Neural Network}
\end{abstract}
\section{Introduction}

Nearly 80\% of the data on the Internet are images and videos. Research on analyzing, understanding and mining these multimedia data has drawn a significant amount of attention from both academia and industry. Labeling or classifying videos is one of the most important requirements, and challenges remained to be solved. Google’s YouTube-8M team introduced a large multi-label video classification dataset which composed of nearly 8 million videos-500K hours of video-annotated with a vocabulary of 4800 visual entities\cite{DBLP:journals/corr/Abu-El-HaijaKLN16}. They used the state-of-the-art Inception-v3 network\cite{DBLP:journals/corr/SzegedyVISW15} pre-trained on ImageNet to extract frame features at one-frame-per-second, providing reliable data supporting for large-scale video understanding. The key task is modeling the long sequence of frame features. The popular methods are LSTM (Long Short-Term Memory Networks)\cite{Hochreiter:1997:LSM:265493.264179}, GRU (Gated recurrent units)\cite{DBLP:journals/corr/ChoMGBSB14}, DBoF(Deep Bag of Frame Pooling)\cite{DBLP:journals/corr/Abu-El-HaijaKLN16}, etc.

In this work, a novel deep convolutional graph based frame feature sequence modeling method is proposed, which aims to mine the complex relationship between video frames and shots, and perform hierarchical semantic abstraction across video. Evaluations are made based on the YouTube8M-2018 dataset, which containing about 5 million videos and 3862 labels. We use the provided frame level Inception-v3 features to training the model, the results show that our model out-performs the RNN based methods.

\section{Related works}

Video feature sequence classification is essentially the the task of aggregating video features, that is, to aggregate $N$ $D$-dimensional features into one $D'$-dimensional feature by mining statistical relationships between these $N$ features. The aggregated $D'$-dimensional feature is a highly concentrated embedding, making the classifier easy to mapping the visual embedding space into the label semantic space. It is common using recurrent neural networks, such as LSTM (Long Short-Term Memory Networks)\cite{Hochreiter:1997:LSM:265493.264179}\cite{srivastava2015unsupervised}\cite{yue2015beyond} and GRU (Gated recurrent units)\cite{DBLP:journals/corr/ChoMGBSB14}\cite{ballas2015delving}, both are the state-of-the-art approaches for many sequence modeling tasks. However, the hidden state of RNN is dependent on previous steps, which prevent parallel computations. Moreover, LSTM or GRU use gate to solve RNN gradient vanish problem, but the sigmoid in the gates still cause gradient decay over layers in depth. It has been shown that LSTM has difficulties in converging when sequence length increase\cite{DBLP:journals/corr/abs-1803-04831}.There also exist end-to-end trainable order-less aggregation methods, such as DBoF(Deep Bag of Frame Pooling)\cite{DBLP:journals/corr/Abu-El-HaijaKLN16}.

\section{Proposed Method}

\begin{figure}
\centering
    \includegraphics[width=1.0\linewidth]{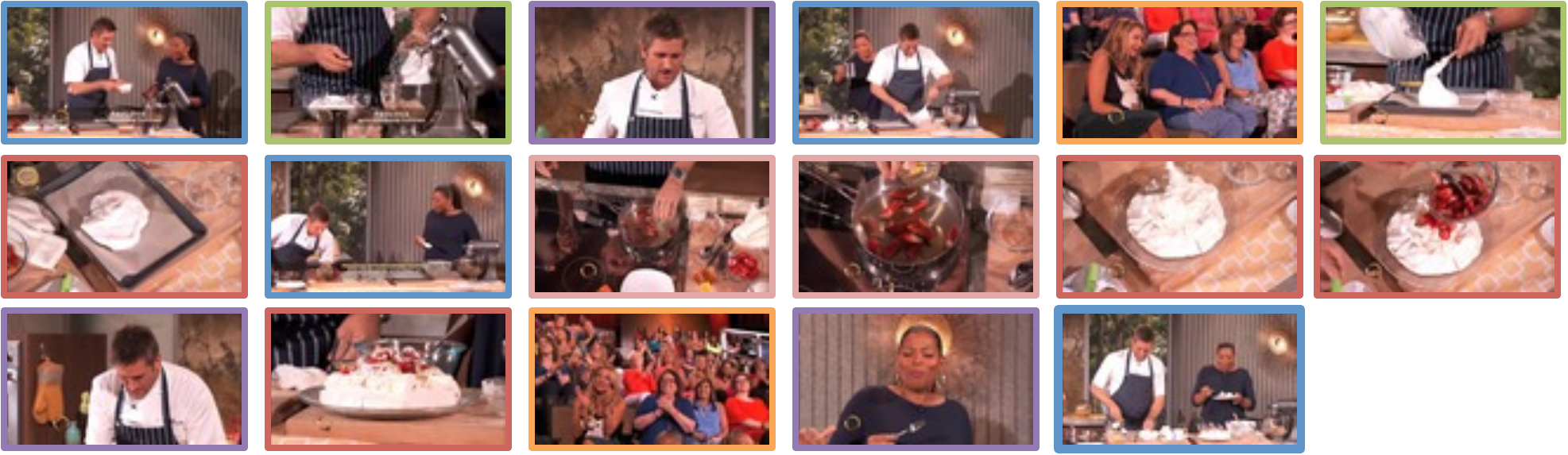}
    \caption{“cooking show” video frame sequence. Frames with same color border contain similar targets.}
    \label{fig:overview_example}
\end{figure}

\begin{figure}
\centering
    \includegraphics[width=1.0\linewidth]{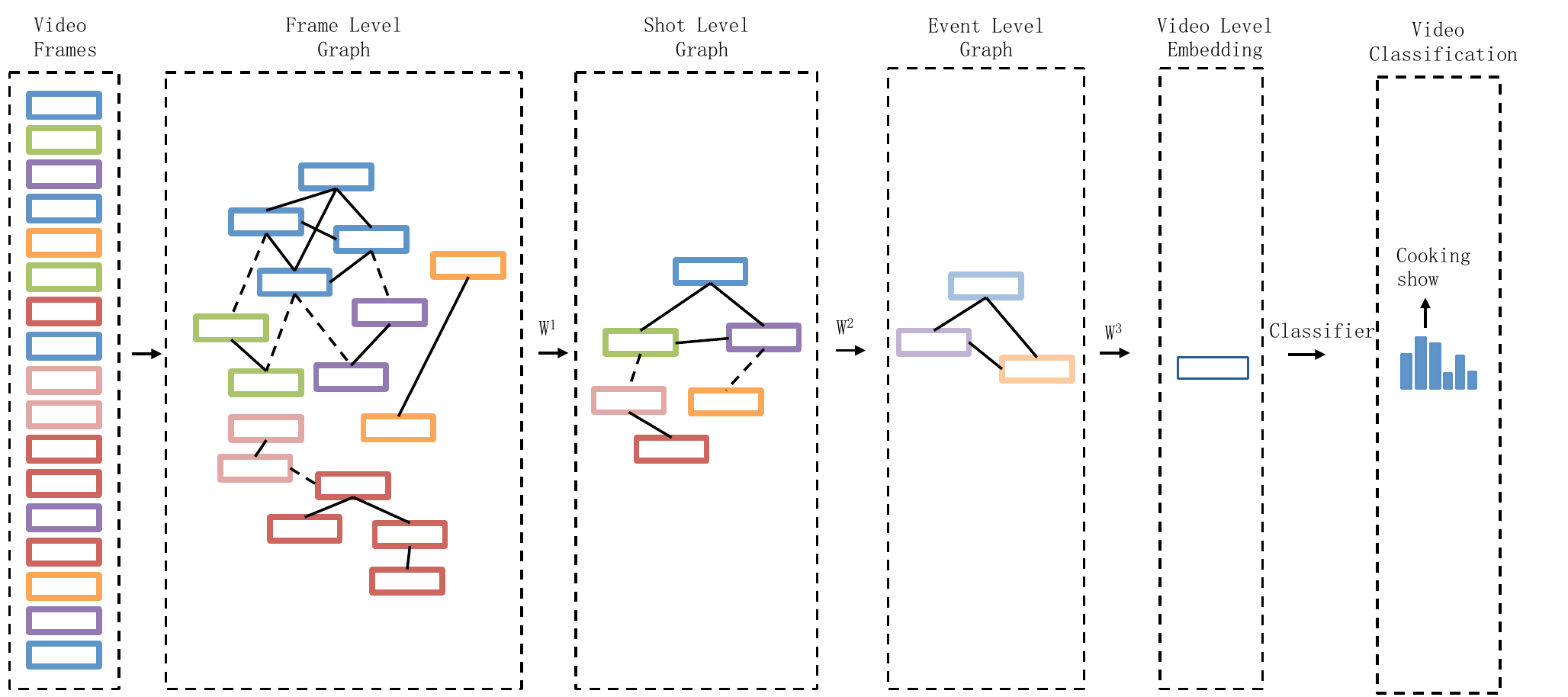}
\caption{High-level illustration of our proposed method DCGN. This is an example with label of "cooking show", containing shots of chef cooking, host chatting, food, audience, etc. Rectangle with different color represent different shots or events. We use graph network to represent the relations between frames, shots, or events, that similar ones(nodes) have edge connected. The graph is gradually aggregated that represent frame, shot, event and video hierarchically. }
\label{fig:overview_example_graph}
\end{figure}

In this paper, we propose a convolutional graph based video representation method for a sequence of video frame features. The main idea is that video is a hierarchical data structure, composed of events, scenes, shots, super-frames and frames. Additionally, the relations between frames, and the relations between shots are more complex than the order of a sequence. Consider an example sequence of “cooking show” as shown in Fig.~\ref{fig:overview_example}, frames containing same targets are not continuous, and distributed in various timestamps, as well as the events and shots. 
We model the video frame sequence, shot, and event hierarchically by a deep convolution graph Network (DCGN). It gradually abstracts information from frame level to video level by convolution and information propagating through graph, and finally generates a global representation for further classification. The method is shown in Fig.~\ref{fig:overview_example_graph}.

\subsection{Graph Network with Deep Convolution}

We use F=$\{f_{i}^{D}, 0\leqslant i\leqslant N\}$ to denote video frame feature sequence, where $D$ is the dimension of the frame feature and $N$ is the number of video frames. We treat each frame, shot or event as a node of a graph, and the graph is densely connected (each pair of the nodes has an edge). Two nodes are connected weighted by their feature vector similarity. Let $A$ be the adjacency matrix of $F$, and the elements is calculated by cosine similarity:
\begin{equation} \label{eq0}
A(i,j)=\frac{\sum_{d=0}^{D-1}(f[i][d]\times f[j][d])}{\sqrt{\sum_{d=0}^{D-1}f[i][d]^{2}}\times \sqrt{\sum_{d=0}^{D-1}f[j][d]^{2}}}
\end{equation}

Graph neural network uses a differentiable aggregation function \ref{eq1} to perform “message passing”. It is an end-to-end learning model, which can learn node and edge representations simultaneously.
\begin{equation} \label{eq1}
h^{l}=G(A,h^{l-1},W^{l})
\end{equation} where $h^{l}$ is the node representation(messages) after $l$-th iteration, $W^{l}$ is the parameters for the  $l$-th iteration. $G$ is the message propagation function. GCN\cite{kipf2016semi} is a popular graph model whose $G$ is:

\begin{equation} \label{eq2}
G=\sigma (\sqrt{\overline{D}}A\sqrt{\overline{D}}h^{l-1}W^{l})
\end{equation}
where $\overline{D}$ is the diagonal node degree matrix that normalizing A such that all rows sum to one.

\subsubsection{Graph pooling.} 
In the equation \ref{eq2} the adjacency matrix $A$ is unchanged during iteration, so the topology of the graph is static. However, video frame sequence is a hierarchical structure, hence the graph topology should be abstracted to higher level gradually. We use two pooling methods to aggregate graph.

\paragraph{average pooling.}
This method applies the center of $K$ consecutive nodes as the node of the next level graph:
\begin{equation}
p^{l}[i][d]=\frac{\sum_{k=0}^{K-1}h^{l-1}[i\times K+k][d]}{K}, d\in [0,D]
\end{equation} where $K$ is the pooling kernel size, $p^{l}$ is the $l$-th pooled graph node feature vector. After $l$-th iteration, the graph size is $\frac{1}{K^{l}}$ of the original. 

\paragraph{self-attention based pooling.}
This method performs a local self-attention to obtain a weight $\alpha$ for each feature of the local consecutive sequence, thereby obtaining a locally weighted and fused output of the feature sequence. Comparing to average pooling, it can better obtain the topology of the next layer graph, which is beneficial to the propagation of feature information. We formulate it as:

\begin{equation}
\begin{split}
p_{att}^{l}[i]=\alpha^{l}[i] \otimes h^{l-1}[i\times K:(i+1)\times K], \\
\alpha^{l}[i]=softmax(h^{l-1}[i\times K:(i+1)\times K]W_{att}^{l}+b^{l})
\end{split}
\end{equation} where $K$ is the number of local features to perform self-attention, and $W_{att}$ and $b$ are the parameters to learn.

\subsubsection{Nodes convolution.} 
Different from GCN\cite{kipf2016semi} which use fully connected layers to represent the nodes, in order to represent frame sequence hierarchically and maintain the local sequence order, we represent the nodes by convolution as follows:
\begin{equation}
c^{l}[i]=h^{l-1}[i\times K:(i+1)\times K]W^{l}
\end{equation}
where $c^{l}$ is the $l$-th graph node embedding, and $W^{l}$ is the convolution kernel weights with size of $K\times D$. 

\begin{figure}
\centering
\includegraphics[width=1\linewidth]{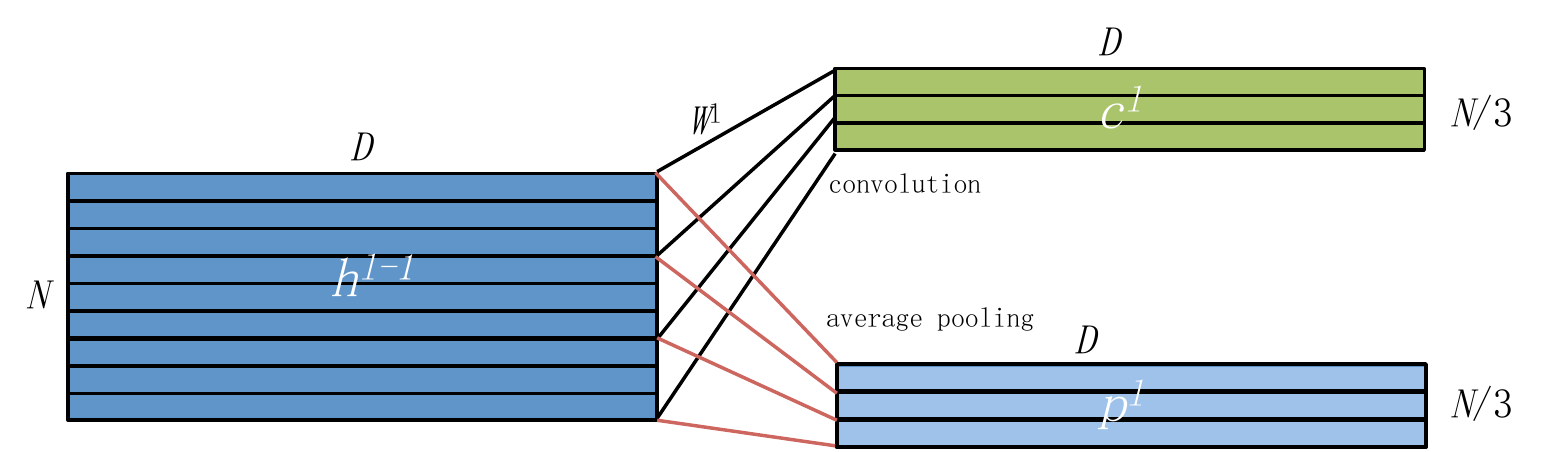}
\includegraphics[width=1\linewidth]{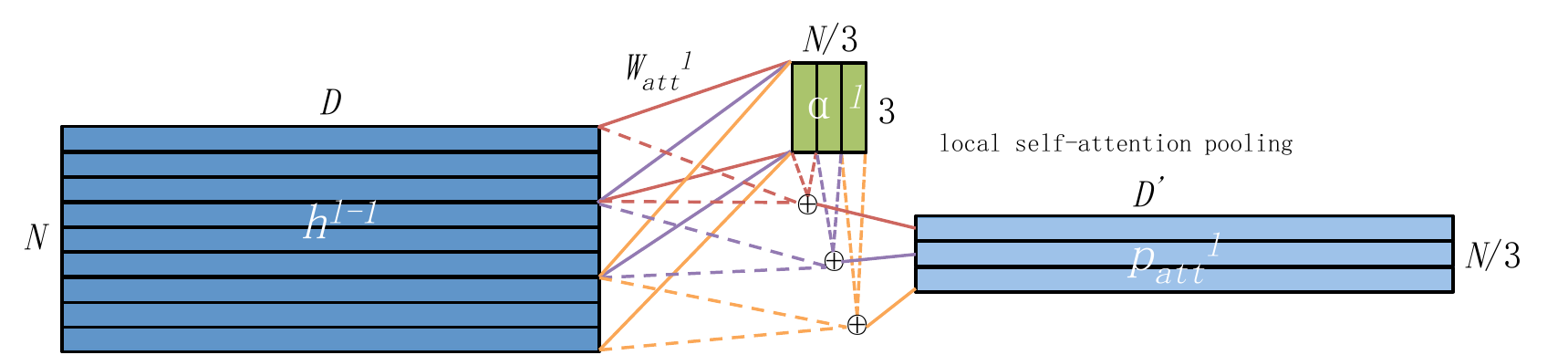}
\caption{Convolution and pooling ($K$=3) of $N$ nodes with $D$-dimension, outputs new $\frac{N}{3}$ nodes with $D$-dimension.}
\label{fig:graph_pool_conv}
\end{figure}

Pooling and convolution are shown in Fig.~\ref{fig:graph_pool_conv}.

\subsubsection{Node feature propagation.}
\begin{figure}
\centering
\includegraphics[width=1\linewidth]{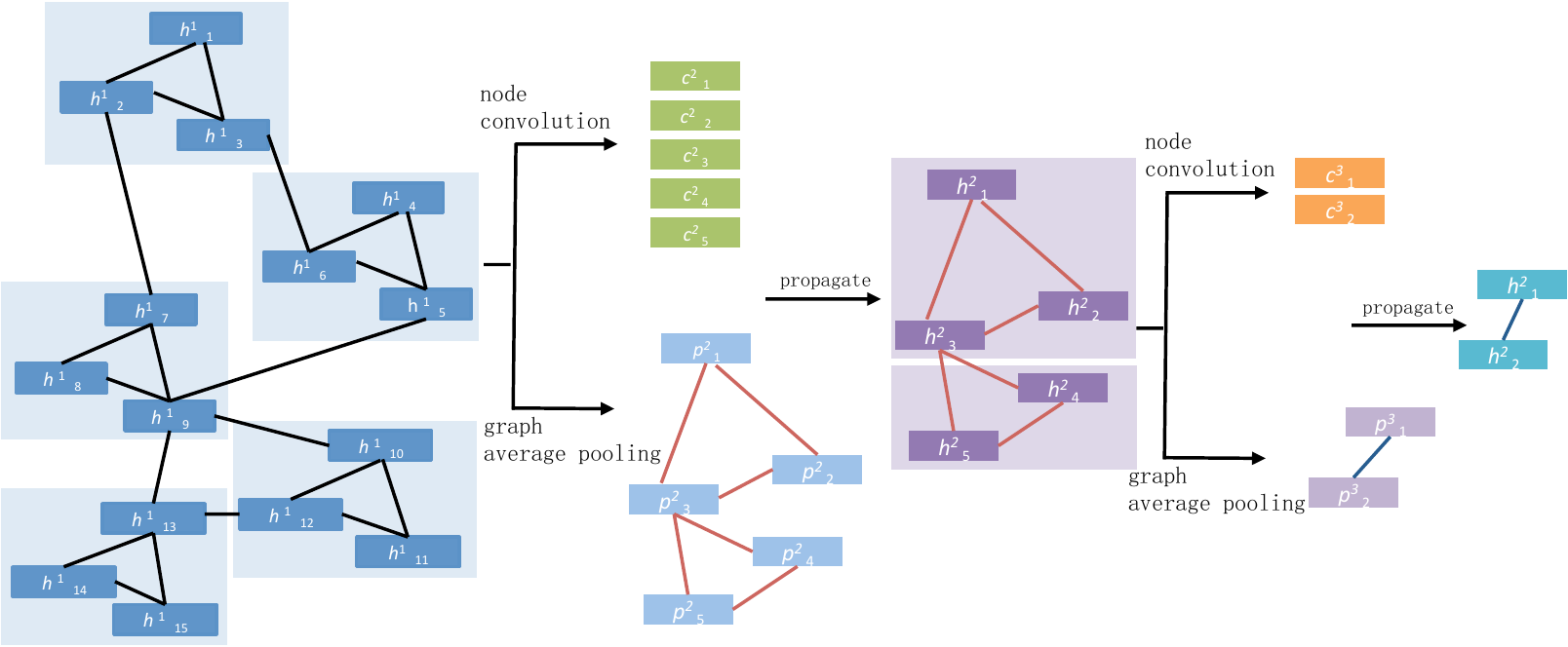}
\caption{Illustration of tow layer convolutional graph network. 15 input feature vectors are aggregated to 2 output feature vectors.}
\label{fig:graph_network}
\end{figure}

Now, we have obtained a new graph topology and new feature vector for each node. In order to obtain a more complete representation in higher level, we perform the “message passing” across the entire graph, so that the fused feature of each node is generated from the global perspective. We use the similar form as equation \ref{eq2}:

\begin{equation}
h^{l}=\sigma (\sqrt{\overline{D}^{l-1}}A^{l-1}\sqrt{\overline{D}^{l-1}}c^{l-1}W^{l})
\end{equation}
where $A^{l-1}$ is calculated using equation \ref{eq0} in which $f$ is replaced with $p^{l-1}$.
Fig.~\ref{fig:graph_network}  shows the network described above.

\subsection{Shot segmentation aided graph pooling}

\begin{figure}
\centering
\includegraphics[width=1\linewidth]{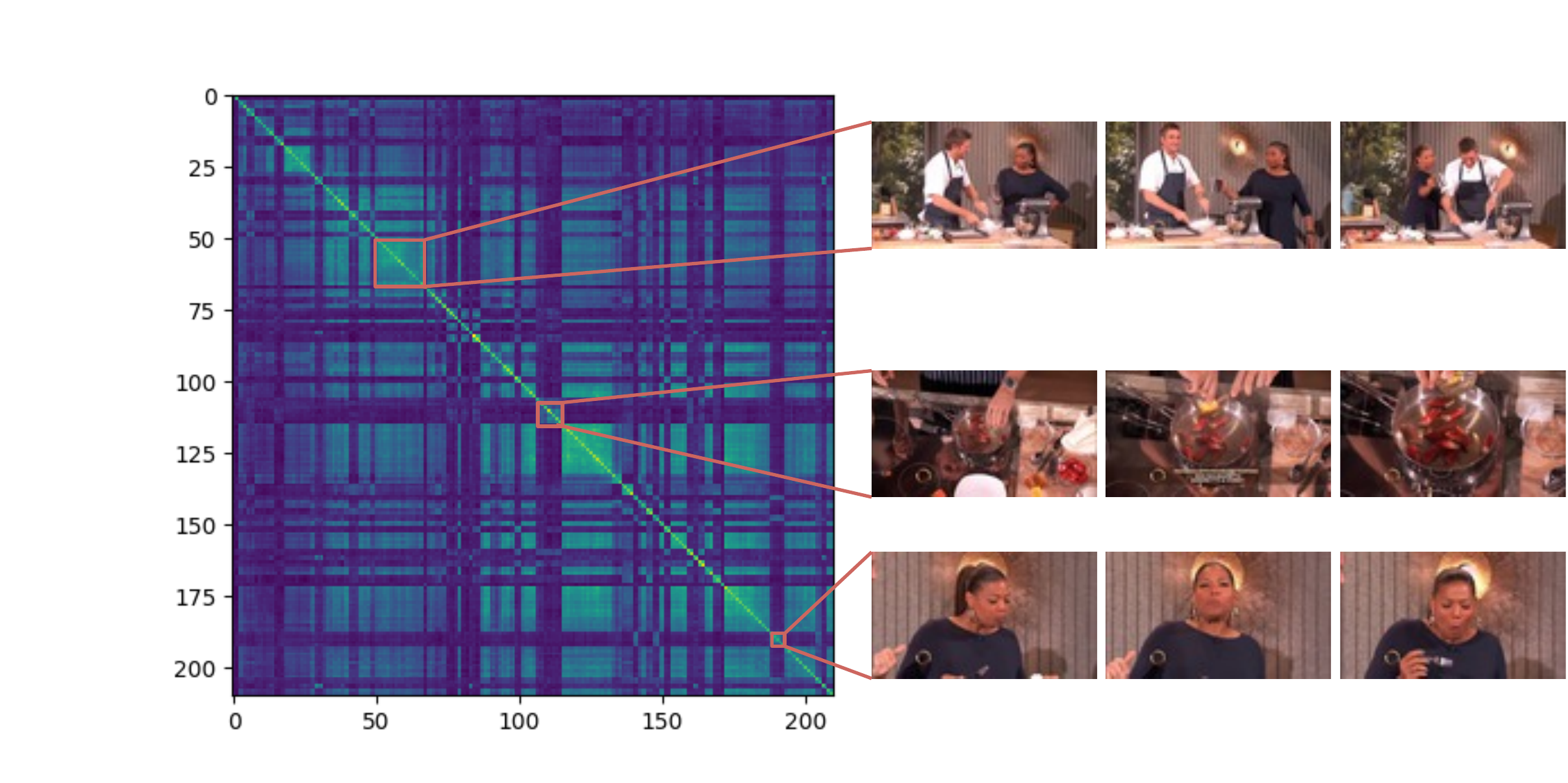}
\caption{The left is a matrix of frame-to-frame similarities, which brighter(larger value) points indicate more similar frames.  Local bright squares are treated as a shot.}
\label{fig:shot_seg}
\end{figure}

Shot is a basic temporal unit, which is a series of interrelated consecutive pictures taken contiguously by a single camera and representing a continuous action in time and space. It is independent over video category, and can be obtained by unsupervised nonparametric way. Inspired by the kernel temporal segmentation (KTS) algorithm \cite{potapov2014category}, the proposed method applies the deep convolutional neural network features to calculate the matrix of frame-to-frame similarities. The algorithm uses dynamic programming to minimize within segment kernel variances, and get the best shot boundaries under given number of shots, the object function is:
\begin{equation}
\min_{m;t_{0},...,t_{m-1}}J_{m,n}:=L_{m,n}+Cg(m,n)
\end{equation}
where
\begin{equation}
\begin{split}
L_{m,n}=\sum_{i=0}^{m}v_{t_{i-1},t_{i}}, 
g(m, n) = m(log(\frac{n}{m}) + 1),\\
v_{t_{i-1},t_{i}}=\sum_{t=t_{i}}^{t_{i=1}-1}\left \|f_{t}-\mu _{i} \right \|^{2},
u_{i}=\frac{\sum_{t=t_{t_{i}}}^{t_{i+1}-1}f_{t}}{t_{i+1}-t_{i}}
\end{split}
\end{equation} where $m$ is the number of shots and $g(m, n)$ is a penalty term. Fig.~\ref{fig:shot_seg}  shows the segmented shots together with their positions in frame-to-frame similarity matrix.

We apply this shot segmentation algorithm to the frame level graph modeling, and the formulation becomes:

\begin{equation}
p^{1}[t][d]=\frac{\sum_{k=s_{t}}^{s_{t+1}}f[k][d]}{s_{t+1}-s_{t}}, d\in[0,D]
\end{equation}

\begin{equation}
c^{1}[t]=f[s_{t}:s_{t+1}]W^{1}
\end{equation} where {t} is the shot number, $s_{t}$ is frame index  of the $t$-th shot boundary. 

\begin{figure}
\centering
\includegraphics[width=1\linewidth]{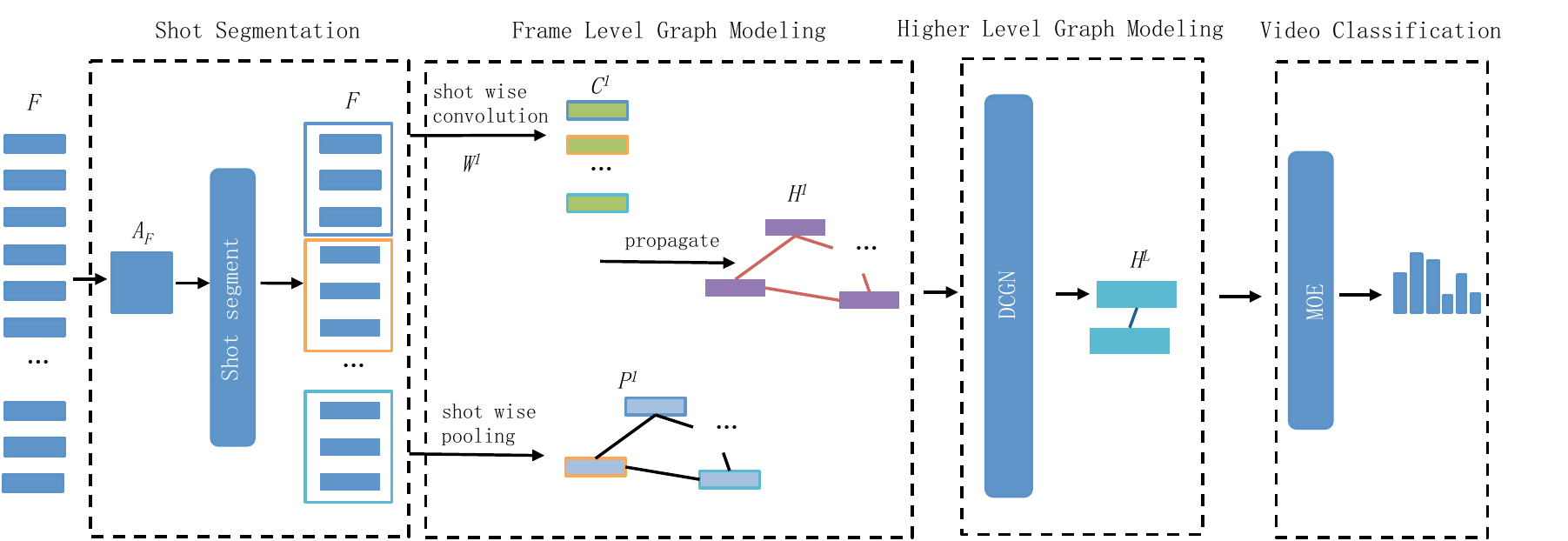}
\caption{Overall architecture. Shot-wise convolution and pooling performs at the frame level.  The subsequent levels perform DCGN and obtain a video level feature $H^{L}$.  MoE classifier use $H^{L}$  as the input to provide the final label predictions.}
\label{fig:shotseg_graph_network}
\end{figure}

Fig.~\ref{fig:shotseg_graph_network} shows the network aided with shot segmentation: first, a fixed number {m} of shots are segmented, and a frame level convolution graph network is applied to get the shot level features $H^{1}$. Then deep convolution graph network is applied to modeling higher-level features. Finally, the last level feature vectors are concatenated and inputted to a mixture of experts (MoE) \cite{jordan1994hierarchical} model to get the classification score. The model is trained end-to-end.

\section{Experiment}

In this section, we evaluate the proposed method with other alternatives, including average pooling, LSTM, GRU, DBoF. We train these models on YouTube-8m-2018 frame-level training set and evaluate on validation set. There are 3.9 million videos in the training set and 1.1 million videos in the validation set. The dataset is a multi-class task with 3862 classes in total, and we use cross entropy loss to train the model:
\begin{equation}
Loss={-}\sum_{c=1}^{C}y_{c}log(p_{c})
\end{equation}  where $C$ is the number of classes, $p_{c}$ is the prediction for class $c$.

We use GAP and Hit@1 \cite{DBLP:journals/corr/Abu-El-HaijaKLN16}\cite{startcode} as the evaluation metrics:

\begin{equation}
GAP=\sum_{i=1}^{N}p(i)\Delta r(i)
\end{equation} where $N$ is the number of final predictions, we set $N$=20 here, $p(i)$ is the precision, and $r(i)$ is the recall.  

\begin{equation}
Hit@1=\frac{1}{|V|}\sum_{v\in V }\vee _{e\in G_{v}}\mathbb{I}(rank_{v,e}\leqslant 1)
\end{equation} where  $G_{v}$ is the set of ground-truth entities for $v$. This is the fraction of test samples that contain at least one of the ground truth labels in the best prediction.

The network architecture parameters and hyper-parameters for training these models are given in Table~\ref{table:trainparams}.

\begin{table}[]
\begin{center}
\caption{Training settings}
\label{table:trainparams}
\resizebox{\textwidth}{17mm}{
\begin{tabular}{lcllll}
\hline
             & \multicolumn{1}{l}{Ours~~~~~~}      & \begin{tabular}[c]{@{}l@{}}Average\\ Pooling~~~~~~  \end{tabular}     & LSTM~~~~~~              & GRU~~~~~~         & DBoF~~~~~~         \\ \hline
network parameters           &  \begin{tabular}[c]{@{}l@{}}layers: 5,~~~~~\\ filter size: 1024~~~~~ \\(same for each layer) \end{tabular}  &                 & \begin{tabular}[c]{@{}l@{}}cells: 1024,~~~~~\\ layers: 2~~~~~\end{tabular} & \begin{tabular}[c]{@{}l@{}}cells: 1024,~~~~~\\ layers: 2~~~~~\end{tabular} & \begin{tabular}[c]{@{}l@{}}cluster size: 8192,\\ hidden size: 1024,\\ pooling method: max\end{tabular} \\ \hline
base learning rate           & \multicolumn{5}{c}{0.001}      \\ \hline
learning rate decay          & \multicolumn{5}{c}{0.8}       \\ \hline
learning rate decay examples & \multicolumn{5}{c}{4M}      \\ \hline   
mini batch size & \multicolumn{5}{c}{1024} \\ \hline  
number of epochs &     \multicolumn{5}{c}{5} \\ \hline
number of mixtures in MoE  &   \multicolumn{5}{c}{2}                                                                                                                                                                                                                                                                                                                                                                                                                                                                                                         \\ \hline
\end{tabular}}
\end{center}
\end{table}

Table ~\ref{table:trainresults} shows results for all methods. Simple averaging of features across all frames, perform poorly on this dataset. The  order-less aggregation methods DBoF perform worse than the RNN based models. Our model outperforms all other models, and the self-attention based graph pooling got the best result.
\begin{table}[]
\begin{center}
\caption{evaluation results. All values in this table are averaged results and reported in percentage(\%). Top-2 of each metrics are bolded.}
\label{table:trainresults}
    \begin{tabular}{llll}
    \hline
    methods~~~                               & GAP~~~   & Hit@1~~~   & Loss     \\ \hline
    Average pooling~~~                   & 76.5~~~    & 83.5~~~           & 5.49    \\ \hline
    LSTM ~~~                                 & 83.6~~~   & 87.2~~~           & 4.12 \\ \hline
    GRU ~~~                                  & 83.9~~~    & {\bf87.9} ~~~          & 4.03    \\ \hline
    DBoF  ~~~                                & 81.1~~~   & 86.2~~~           & 6.02 \\ \hline
    ours + average graph pooling~~~          & {\bf84.1} ~~~  & 87.4         & {\bf4.01}   \\ \hline
    ours + self-attention graph pooling~~~   & {\bf84.5} & {\bf87.7}~~~       & {\bf3.98}      \\
    \end{tabular}
    \end{center}
\end{table}

\section{Conclusions}

In this work, we propose a novel sequence representation method DCGN for video classification. One layer in DCGN is composed of graph pooling, nodes convolution and nodes feature propagation. The problem of representing complex relationships between frames or shots is addressed by applying this graph based network hierarchically across the frame sequence. Based on the quantitative results in Section 4, the proposed method DCGN outperform the other alternatives such as LSTM and GRU.

%
%
%

\begin{thebibliography}{10}

\bibitem{DBLP:journals/corr/Abu-El-HaijaKLN16}
Abu{-}El{-}Haija, S., Kothari, N., Lee, J., Natsev, P., Toderici, G.,
  Varadarajan, B., Vijayanarasimhan, S.:
\newblock Youtube-8m: {A} large-scale video classification benchmark.
\newblock CoRR \textbf{abs/1609.08675} (2016)

\bibitem{DBLP:journals/corr/SzegedyVISW15}
Szegedy, C., Vanhoucke, V., Ioffe, S., Shlens, J., Wojna, Z.:
\newblock Rethinking the inception architecture for computer vision.
\newblock CoRR \textbf{abs/1512.00567} (2015)

\bibitem{Hochreiter:1997:LSM:265493.264179}
Hochreiter, S., Schmidhuber, J.:
\newblock Long short-term memory.
\newblock Neural Comput. \textbf{9}(9) (November 1997)  1735--1780

\bibitem{DBLP:journals/corr/ChoMGBSB14}
Cho, K., van Merrienboer, B., G{\"{u}}l{\c{c}}ehre, {\c{C}}., Bougares, F.,
  Schwenk, H., Bengio, Y.:
\newblock Learning phrase representations using {RNN} encoder-decoder for
  statistical machine translation.
\newblock CoRR \textbf{abs/1406.1078} (2014)

\bibitem{srivastava2015unsupervised}
Srivastava, N., Mansimov, E., Salakhudinov, R.:
\newblock Unsupervised learning of video representations using lstms.
\newblock In: International conference on machine learning. (2015)  843--852

\bibitem{yue2015beyond}
Yue-Hei~Ng, J., Hausknecht, M., Vijayanarasimhan, S., Vinyals, O., Monga, R.,
  Toderici, G.:
\newblock Beyond short snippets: Deep networks for video classification.
\newblock In: Proceedings of the IEEE conference on computer vision and pattern
  recognition. (2015)  4694--4702

\bibitem{ballas2015delving}
Ballas, N., Yao, L., Pal, C., Courville, A.:
\newblock Delving deeper into convolutional networks for learning video
  representations.
\newblock arXiv preprint arXiv:1511.06432 (2015)

\bibitem{DBLP:journals/corr/abs-1803-04831}
Li, S., Li, W., Cook, C., Zhu, C., Gao, Y.:
\newblock Independently recurrent neural network (indrnn): Building {A} longer
  and deeper {RNN}.
\newblock CoRR \textbf{abs/1803.04831} (2018)

\bibitem{kipf2016semi}
Kipf, T.N., Welling, M.:
\newblock Semi-supervised classification with graph convolutional networks.
\newblock CoRR \textbf{abs/1609.02907} (2016)

\bibitem{potapov2014category}
Potapov, D., Douze, M., Harchaoui, Z., Schmid, C.:
\newblock Category-specific video summarization.
\newblock In: European conference on computer vision, Springer (2014)  540--555

\bibitem{jordan1994hierarchical}
Jordan, M.I., Jacobs, R.A.:
\newblock Hierarchical mixtures of experts and the em algorithm.
\newblock Neural computation \textbf{6}(2) (1994)  181--214

\bibitem{startcode}
Google:
\newblock Youtube-8m starter code.
\newblock \url{https://github.com/google/youtube-8m} 2018.

\end{thebibliography}
%

\end{document}